\documentclass{article}
\usepackage{ijcai24}

\usepackage{times}
\usepackage{soul}
\usepackage{url}
\usepackage[hidelinks]{hyperref}
\usepackage[utf8]{inputenc}
\usepackage[small]{caption}
\usepackage{graphicx}
\usepackage{amsmath}
\usepackage{amsthm}
\usepackage{booktabs}
\usepackage{algorithm}
\usepackage[switch]{lineno}

\usepackage{url}
\usepackage{stfloats}
\usepackage{amsfonts}
\usepackage{subfig}
\usepackage{bm}
\usepackage{bbm, dsfont}
\usepackage{algpseudocode} 
\usepackage{multirow}
\usepackage{array}
\usepackage{makecell}
\usepackage{adjustbox}

\title{Auxiliary Reward Generation with Transition Distance Representation Learning}

\author{
Siyuan Li$^1$
\and
Shijie Han$^2$\and
Yingnan Zhao$^{2,3}$\and
By Liang $^4$\And
Peng Liu $^5$\\
\affiliations
$^1$Harbin Institute of Technology\\
$^2$Harbin Institute of Technology\\
$^3$Harbin Engineering University \\
$^4$CALT\\
$^5$Harbin Institute of Technology\\
}

\begin{document}

\maketitle

\begin{abstract}
    Reinforcement learning (RL) has shown its strength in challenging sequential decision-making problems. The reward function in RL is crucial to the learning performance, as it serves as a measure of the task completion degree. In real-world problems, the rewards are predominantly human-designed, which requires laborious tuning, and is easily affected by human cognitive biases. To achieve automatic auxiliary reward generation, we propose a novel representation learning approach that can measure the ``transition distance'' between states. Building upon these representations, we introduce an auxiliary reward generation technique for both single-task and skill-chaining scenarios without the need for human knowledge. The proposed approach is evaluated in a wide range of manipulation tasks. The experiment results demonstrate the effectiveness of measuring the transition distance between states and the induced improvement by auxiliary rewards, which not only promotes better learning efficiency but also increases convergent stability.
\end{abstract}

\section{Introduction}
Reinforcement learning (RL) \cite{sutton2018reinforcement} is a promising paradigm for solving 
 sequential decision-making problems, where an agent tries to learn policies by trial and error. The reward function is crucial to RL, which significantly influences learning efficiency and convergent policies, and an effective reward function should accurately describe the extent of task completion. In real-world problems, the reward functions are mostly designed by human experts, which require much labor for tuning and are heavily influenced by human cognitive biases. Due to the importance and challenges of designing rewards, automatic reward generation has received much attention from researchers.


\begin{figure}[htbp]
\centering
    \subfloat[]{
        \includegraphics[width=0.18\textwidth]{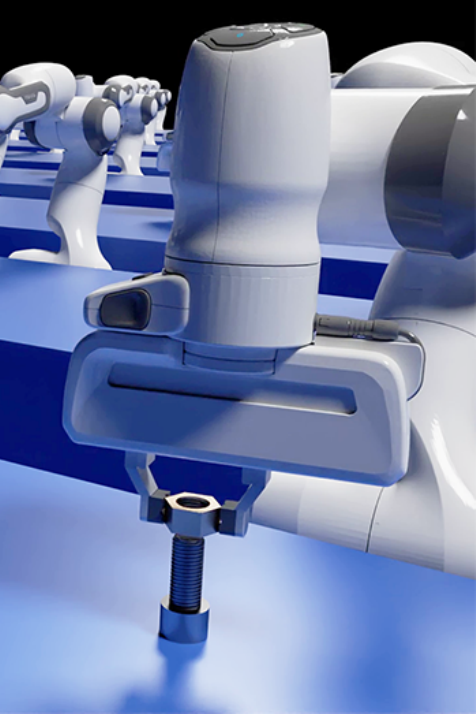}
        \label{fig1:a}}
    \subfloat[]{
        \includegraphics[width=0.22\textwidth]{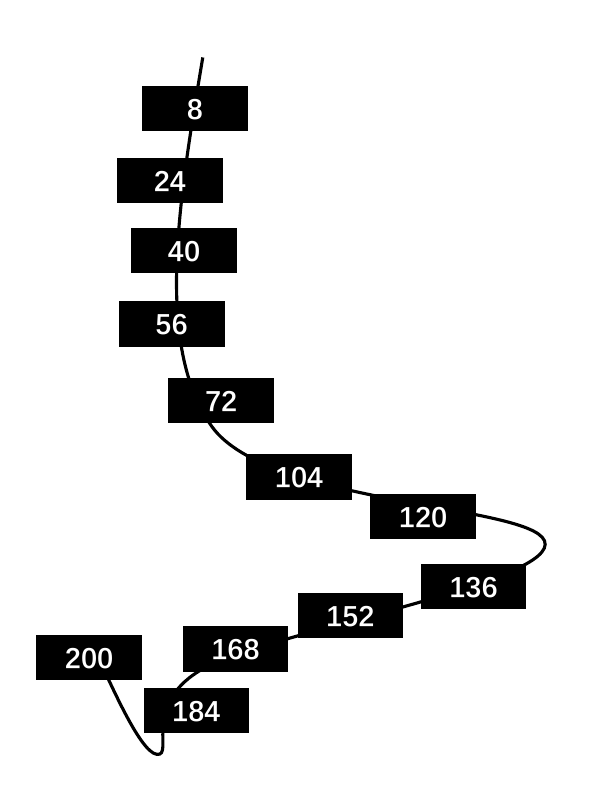}
        \label{fig1:b}}
\caption{Euclidean distance in the raw state space cannot measure the difficulty of achieving state transitions. (a) The Place-nut task. (b) A  successful trajectory in the x-z space, where a square represents a state, and the number denotes the timestep index of the state.}
\label{fig1}
\end{figure}

 Two primary categories of reward generation approaches are reward shaping and curriculum learning. 
 Reward shaping, such as the potential-based reward shaping \cite{ng1999policy} and more recent methods \cite{2018Belief,li2019hierarchical,2019Reward,2020Learning}, involves modifying the original reward function to integrate domain knowledge into RL. 
From a different view, curriculum RL \cite{2009Curriculum} organizes policy learning in a task sequence, from easier to more difficult ones.
The automated task generation in curriculum RL is closely related to reward generation \cite{2016Source,2018Object}, as in the tasks formulated as Markov decision processes (MDPs), the rewards are of great importance.
Most curriculum RL works generate rewards based on the context variables in the contextual MDPs \cite{hallak2015contextual}.
Although there has been progress in automatic reward generation, existing methods struggle to precisely measure how close the agent is to completing the whole task from the current state, which is critical to effective reward design.  


As most task completion can be described with a goal state, an intuitive idea of measuring task completion degree is by the distance between the current state and the goal state.
However, a crucial problem is how to define the distance. In real-world problems, such as robotics manipulation and maze navigation, the Euclidean distance in the state space makes no sense. For example, Figure \ref{fig1}(b) represents a trajectory in the Place-nut task in the x-z space, and the number denotes the timestep index of the corresponding state in an episode. As shown by Figure \ref{fig1}(b), state 136 is closer to completing the task than state 104, however, the Euclidean distance between state 104 and the goal state 200 is less than the distance between state 136 and the goal state 200.
As the distance in the raw space cannot measure the task completion degree, in this paper, we propose a novel approach to generate auxiliary rewards based on Transition Distance RePresentation (TDRP) - that measures the ``transition distance" between two states. Transition distance quantifies the number of states within a transition sequence required to move from one state to another, which is proportional to the degree of task completion. In the representation space abstracted by TDRP, the Euclidean distance between two representation states can represent the length of the transition distance to some extent. Specifically, a smaller distance between two representation states implies that only a few actions are required to transition from one state to another. With this representation, we develop goal-based dense auxiliary rewards for single tasks and long-horizon skill-chaining tasks.

We evaluate the efficacy of the proposed TDRP approach in a wide range of robot manipulation tasks and demonstrate that the distance between the embeddings learned by TDRP can measure the transition distance. The experiment results show that the auxiliary rewards generated by our method can significantly improve the sample efficiency and the convergent performance for single tasks and skill-chaining tasks. Besides, the experiment results indicate that our auxiliary reward method produced by TDRP significantly outperforms the state-of-the-art representation learning methods and reward-shaping methods in RL.

\section{Preliminaries}\label{section3}
The learning environment of RL is formulated as an MDP, described by a tuple $M=(S,A,P,R,\gamma)$. Here, $S$ is the state space, and $A$ is the action space. 
$P:S\times A \times S \rightarrow \mathbb{R}$ is 
the transition probability, and $r=R(s,a)$ is a reward function, $\gamma \in [0,1)$ is a discount factor. 
At timestep $t$, an agent at state $s_t\in S$ executes the action $a_t \in A$ specified by the policy $\pi_{\theta}: a_t\sim\pi_{\theta}(s_t)$, and transits to the  next state $s_{t+1}$  according to the transition probability $s_{t+1}\sim P(s_t,a_t)$, and receives a reward $r_t=R(s_t,a_t)$. We use $\tau_i=\{(s_0,a_0),(s_1,a_1)...(s_T,a_T)\}$ to represent a trajectory of length $T$ form trajectory set: $\tau_i\in\bm{\tau}$. The value function $V_{\alpha}(s)$ is updated as Equation \eqref{equl0.0}. 
\begin{equation}
    \arg\min_{\alpha}\frac{1}{|\bm{\tau}|T}\sum_{\tau_i\in\bm{\tau}}\sum_{t=0}^{T} 
    (V_{\alpha} (s_t)-(\sum_{t^{'}=t}^{\infty}(\gamma^{t^{'}-t}r_{t^{'}})))^2
    \label{equl0.0}
\end{equation}
The advantage function is defined as $A^{\pi}(s_t,a_t)=-V(s_t)+\sum_{t^{'}=t}^{t+H-1}(y^{t^{'}-t}r_{t^{'}})+y^{H}V(s_H)$, $H$ is a length which is much less than the episode length $T$. The policy $\pi_{\theta}$ is updated with the data from the old policy $\pi_{\theta_{k}}$ as Equation \eqref{equl0.1}, and $\epsilon$ is a clipping parameter.
\begin{equation}
\begin{aligned}
    &\arg\max_{\theta}E_{s,a\sim\pi_{\theta_{k}}}[L(s,a,\theta_{k},\theta)] \\
    &\begin{aligned}
        L(s,a,\theta_{k},\theta)=&min(\frac{\pi_{\theta}(a|s)}{\pi_{\theta_{k}}(a|s)}A^{\pi_{\theta_{k}}}(s,a), \\ 
        &clip(\frac{\pi_{\theta}(a|s)}{\pi_{\theta_{k}}(a|s)},1-\epsilon, 1+\epsilon)A^{\pi_{\theta_{k}}}(s,a))
    \end{aligned}
    \label{equl0.1}
\end{aligned}
\end{equation}

\begin{figure*}[htbp]
\centering
\includegraphics[width=0.8\linewidth]{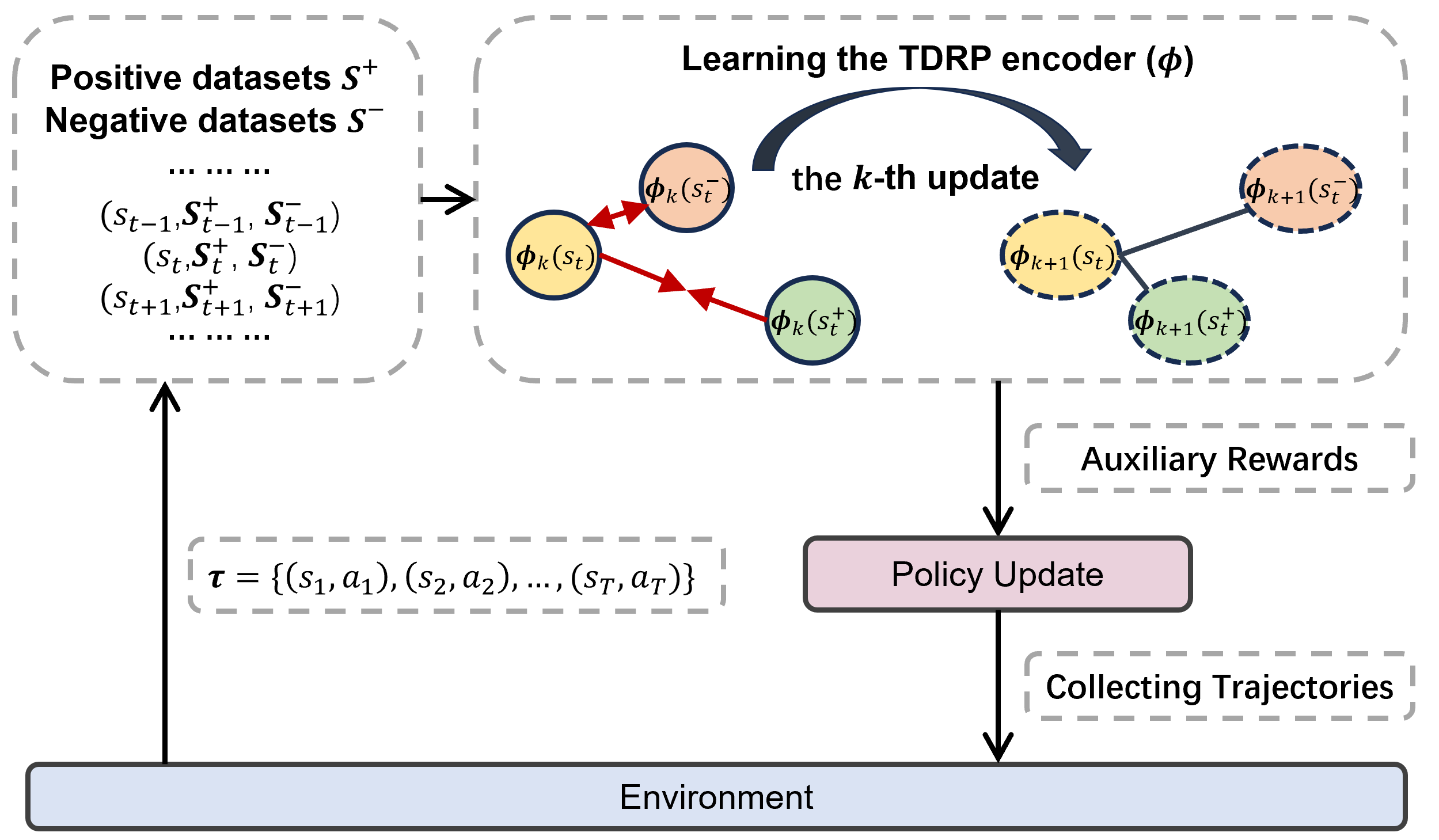}
\caption{The proposed learning framework, where the TDRP model and the policy are simultaneously learned.}
\label{fig2}
\end{figure*}

\section{Method}
To generate distance-based auxiliary rewards, a primary question is how to learn a proper representation space where the Euclidean distance can measure the extent to which the state is distant from task completion. This section first presents the proposed transition distance representation learning approach, and then elaborates how we use this representation to generate auxiliary rewards. Finally, we present the pseudo-code to clearly present all the procedures.

\subsection{Transition Distance Representation Learning}\label{section4}
The core of TDRP is that the distance between the latent embeddings can measure the transition numbers between two states.
This requires the representation to eliminate irrelevant information that has no bearing on task execution. High-dimensional observations typically contain a great deal of sensor signal information, such as camera images, temperature, and force sensors. However, most information in observation is irrelevant to the task at hand, and directly calculating Euclidean distance between raw states cannot accurately represent their timestep index in the trajectory distribution. To address this issue, we propose TDRP for measuring transition distance. We expect the distance between two embeddings encoded by the TDRP model
 proportional to the number of transitions between the two given states. For instance, if $s_i$ and $s_j$ are states in timesteps $i$ and $j$, respectively, the distance between their TDRP-encoded representation can be expressed as 
\begin{equation}
||\phi(s_i)-\phi(s_j)||_2\propto|j-i|.
\end{equation}
To achieve this goal, TDRP must focus on extracting factors that are most relevant to decision-making and state transitions while discarding noisy information that is unrelated to the task control. Reconstruction-based methods cannot achieve this goal because even irrelevant information is necessary to reconstruct the state. Therefore, we propose to achieve TDRP learning based on the contrastive objective.

The goal of contrastive representation learning is to minimize the distance between similar data points while maximizing the distance between dissimilar points.
The key in contrastive learning is to define the similarity, i.e., the positive pairs and the negative pairs.
As we aim to learn a representation space where the distance can describe the transition number, we propose to use the transition number between two states in a trajectory to determine their similarity 
As shown in Figure \ref{fig2}, for state $s_t \in S$ in each trajectory in $\bm{\tau}$, we collect the positive dataset $\textbf{S}_t^+$  and negative dataset $\textbf{S}_t^-$ of the state as 
\begin{align}
\textbf{S}_t^+&=\{s_{t+1},s_{t+2},\dots,s_{t+step}\}, \\  
\textbf{S}_t^-&=\{s_{t+step+1},s_{t+step+2},\dots,s_{t+2*step}\},
\end{align}
where ``$step$" is a hyper-parameter, used to control the size of the positive sample set and negative sample set. These datasets are used to train the TDRP model with the loss function in Equation \eqref{equl2}, which is in a triplet loss form. 
\begin{equation}
    \begin{split}
        Loss &= min_{\phi}E_{\tau_i\in\bm{\tau},s_t\in\tau_i}(\sum_{s_t^+\in \textbf{S}_t^+}||\phi(s_t)-\phi(s_t^+)||_2)+\\&\sum_{s_t^-\in \textbf{S}_t^-}max(1-||\phi(s_t)-\phi(s_t^-)||_2,0))
    \end{split}
    \label{equl2}
\end{equation}
We aim to increase the separation between the representations of states that are far apart in the trajectory while reducing the distance between the representations of states that are adjacent in the trajectory. 

As illustrated in Figure \ref{fig2}, the TDRP encoder can be trained concurrently with the policy learning process. The trajectories collected by the agent are employed to construct the positive datasets and the negative datasets. Simultaneously, the TDRP model provides a representation space to generate auxiliary rewards, which improves the policy learning process. 
Alternatively, the TDRP model can be employed with the pre-collected trajectory datasets as well, such as in the offline RL setting \cite{levine2020offline}.
 
\subsection{Auxiliary Reward Generation}\label{section5}
With the learned TDRP representation, we can set distance-based auxiliary rewards, which indicate the task completion status. In this subsection, we design two kinds of auxiliary rewards, respectively for the single tasks and the long-horizon skill chaining tasks \cite{konidaris2009skill}.

\textbf{Single Tasks.}
We propose an auxiliary reward generation method based on TDRP for a single task. In continuous control systems, goal-reaching tasks often have sparse external rewards, which make policy learning extremely difficult. To overcome this challenge, the TDRP model can be used to generate a dense auxiliary reward for such tasks by calculating the Euclidean distance between the TDRP representation of the current state and the goal state. A small transition distance indicates that reaching the goal from the current state requires only a few actions, resulting in a higher reward. This guides effective exploration and encourages the agent to swiftly approach the goal. For a goal-conditioned task, we set the goal state as $s_g$, the current state as $s$, and use TDRP to develop the auxiliary reward function as Equation \eqref{equl3}.  
\begin{equation}
    r'=R(s,a)-\lambda_1||\phi(s)-\phi(s_g)||_2
    \label{equl3}
\end{equation}
When there are no provided goal states, we acquire task demonstrations to extract several final states from successful trajectories, forming the goal state set. These states are encoded using the TDRP encoder $\phi$ and grouped into $n$ clusters with the k-means method \cite{ahmed2020k}. 
The center point of the $k$-th cluster is denoted as $\phi(s)_{g}^{k}$. We encode the current state $s$ using TDRP and compute the auxiliary reward for the single task as the smallest Euclidean distance between this state representation and the $n$ cluster center points, denoted as Equation \eqref{equl4}.

\begin{equation}
    r'=R(s,a)-\lambda_1\min_{k=1\ to\ n}||\phi(s)-\phi(s)_{g}^{k}||_2
    \label{equl4}
\end{equation}

\textbf{Skill Chaining.}
Skill chaining \cite{konidaris2009skill} involves sequencing pre-learned single skills to complete a related long-horizon task. 
 A major challenge in skill chaining is to ensure the seamless execution of the adjacent skills.
 If the terminal state $s_T^{pre}$ of the previous skill is not included in the initial state set $\textbf{S}_I^{next}$ of the latter skill, it can be difficult to execute the latter skill, resulting in overall task failure. 
 There are typically three methods to address this challenge. The first method involves studying a transition policy \cite{bib12,bib13,bib14} that transits the state from the terminal state of the previous skill to the initial state of the latter skill, thereby requiring the learning of a new policy and incurring significant costs. The second method is to expand the initial state set of the latter skill to include the terminal state set of the previous skill \cite{bib15}, resulting in multiple growths of the state space as the number of skills in the sequence increases. The third method involves constraining the terminal state of the previous skill to the initial state set of the latter skill \cite{bib16}, which can be implemented using TDRP. 
 Specifically, we use the TDRP encoder $\phi$ to encode the states in the initial state set of the latter skill $\textbf{S}_I^{next}$ and partition the representations into $n$ clusters using k-means. The $k$-th representation cluster center point is denoted by $\phi(s)_{I-k}^{next}$. We encode the state of the previous skill using TDRP and select the smallest Euclidean distance between this state's representation and the $n$ representation cluster center points as the auxiliary reward for the previous skill task, denoted as Equation \eqref{equl5}. 
\begin{equation}
    r'=R(s,a)-\lambda_2min_{k=1\ to\ n}||\phi(s)-\phi(s)_{I-k}^{next}||_2
    \label{equl5}
\end{equation}
The auxiliary reward above is utilized to fine-tune the pretrained skills, aiming to lead the agent closer to the initial state of the latter skill, which facilitates the transition between the two skills.

\subsection{Pseudocode} 
We provide the pseudocode in Algorithms 1 to elaborate on the proposed approach.
 The TDRP encoder undergoes simultaneous training with the policy learning process, where the policy is optimized with the Proximal Policy Optimization algorithm (PPO) \cite{schulman2017proximal} in Line 11\footnote{The proposed approach can be applied with other actor-critic policy gradient \cite{konda1999actor} algorithms as well.}.
 Trajectories obtained through trial-and-error in the environment serve as the basis for constructing both positive and negative datasets, subsequently utilized to update the TDRP encoder. Then, we generate auxiliary rewards with the learned representations to achieve better policy learning.

 \floatname{algorithm}{Algorithm}
\begin{algorithm}[ht]
\begin{algorithmic}[1]
\caption{}
\Require {\bf Trajectory buffer: $\tau$, Goal set: $S_g$}
\item Initialize the policy network: $\pi_{\theta}$ and value function $V_{\alpha}$, initialize the TDRP encoder: $\phi$
\For{$i=1, 2, 3$ \ldots $I_{end}$}
\For{$t=0, 1, 2,$ \ldots, $T$}
\State Get $s_t$, $\bm{S}_t^+$, $\bm{S}_t^-$ from $\tau$.
\State Updata $\phi$ with the loss function in Equation \eqref{equl2}.
\EndFor
\State $\phi(s)_g^{1,2\ldots,n} = k$-means$(\phi(S_g))$
\For{$t=0, 1, 2, \ldots, T$}
\State $a_t\sim \pi_{\theta}(a|s_t)$
\State Execute the action $a^t$, get reward $R(s_t, a_t)$
\State Use 
$R(s_t, a_t)$, $\phi$, $(s_t,a_t)$ and $\phi(s)_g^{1,2\ldots,n}$ to generate auxiliary reward $r'$.
\State Update $\pi_{\theta}$ and $V_{\alpha}$ with the new rewards.
\State Store data $(s_t, a_t)$ in $\tau$
\EndFor
\If{Succeed}
    \State Add the final state $s_T$ to the $\textbf{S}_g$ 
    \EndIf
\EndFor
\end{algorithmic}
\label{algo1}
\end{algorithm}

\begin{figure*}[t!]
\centering
    \includegraphics[width=0.99\textwidth]{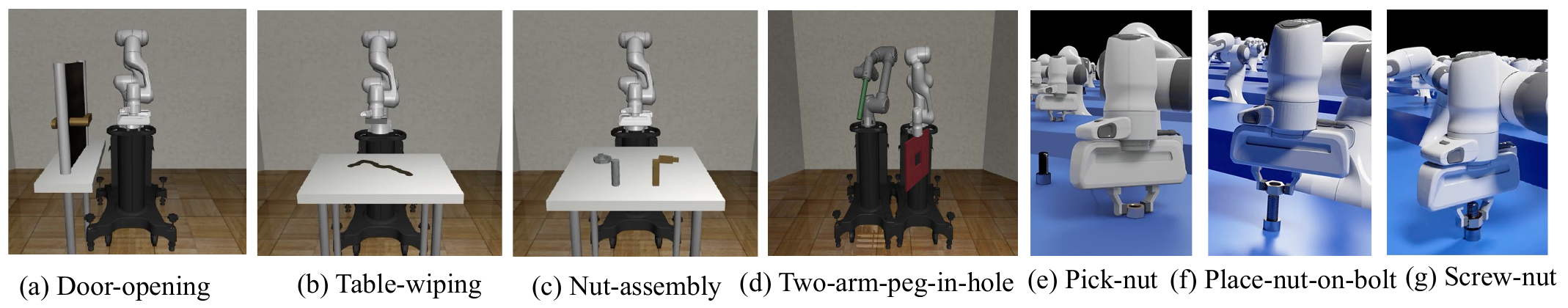}
\caption{Robot manipulation tasks used in the experiment section. Subfigures (a)(b)(c)(d) illustrate the tasks in Robosuite benchmark.
Subfigures (e)(f)(g) illustrate the tasks in the Factory benchmark. }
\label{fig3}
\end{figure*}

 \section{Related Work}\label{section2}
 This section reviews the related work in auxiliary reward generation and representation learning approaches in RL.
\subsection{Auxiliary Rewards}
Auxiliary reward generation has recently received significant attention. Two primary paradigms for generating auxiliary rewards are reward shaping and curriculum learning. Reward shaping incorporates domain knowledge into policy learning by modifying the original reward function. The BiPaRS method \cite{2020Learning} formulates reward shaping as a bi-level optimization problem. The lower level involves optimizing the policy using shaping rewards, while the upper level focuses on optimizing a parameterized shaping weight function for true reward maximization. Additionally, \cite{2019Reward} proposes a meta-learning framework to shape rewards efficiently. Other related methods set auxiliary rewards based on the advantage functions \cite{2018Belief,li2019hierarchical}.
Curriculum RL \cite{2009Curriculum} tackles reward generation from a different view from reward shaping, which gradually learns policies in an automatically generated task sequence with increasing difficulty.
Related to our work, curriculum RL also requires generating reward functions \cite{2016Source,2018Object}, from easier to harder ones. The reward generation in curriculum RL is mostly conducted in a provided context space, such as a goal space or a task-related space.
Although there has been progress in automatic reward generation, the existing works can hardly measure the task completion degree at each state, which is critical to reward generation.

\subsection{Representation Learning}
Existing representation learning approaches in RL can be mainly divided into three categories: reconstruction-based representation, model-based representation, and contrastive representations.

\textbf{Reconstruction-based representations.} Previous methods \cite{bib1,bib2} deal with image observations in two learning phases.
In the first phase, a variational auto-encoder \cite{kingma2013auto} is trained with a reconstruction loss and a KL regularization to transform the image observations into low-dimensional embeddings. In the second phase, policy learning is conducted based on the learned representations.
Although these methods can effectively utilize unlabeled data to train the auto-encoder and improve policy performance, they often extract all factors of the state, including interface information that may be irrelevant to downstream tasks. 
As reconstructing all the information in the observations is not reasonable, the following works \cite{yu2022mask} propose a mask-based reconstruction method with representation prediction, which only reconstructs the information under the mask. 

\textbf{Model-based representations.} 
In model-based RL approaches \cite{bib3,bib4,bib5,bib6}, the encoder and dynamic model are end-to-end trained, leading to more task-oriented representation learning. 
Similar to how the human brain uncovers hidden underlying states, world models \cite{hafner2023mastering,hafner2023mastering}  serve as a natural pre-training objective, enabling the capture of essential environmental parameters. These internal representations of the world significantly impact an agent's decision-making process, influencing the actions deemed likely to yield greater rewards. 
However, the methods with latent model learning can hardly accurately predict trajectories and rewards over long periods. 


\textbf{Contrastive representations.} Contrastive learning \cite{bib11,CL2005,FaceNet2015,InfoNCE2018} aims to learn a representation space where similar states are represented as being close to each other while dissimilar states are far apart. The ``similarity metric" used to measure the similarity between two states is task-specific. Data augmentation is often employed in this method, such as augmenting the same point data to obtain a high similarity \cite{laskin2020curl}, increasing the similarity between adjacent patches \cite{bib10}, and decreasing the similarity between different data points. 

While existing representation learning methods have shown their strength in RL, these methods can hardly fully capture the concept of ``transition distance" - a measure of the complexity of the transition between two states. To address this issue, we propose TDRP, a novel unsupervised representation learning method that can accurately measure transition distance, and then use this measurable representation to generate auxiliary rewards to improve policy learning.

\section{Experiments}\label{section6}
The experiment evaluation is designed to answer the following questions:
\begin{itemize}
    \item Can the auxiliary rewards generated by TDRP facilitate faster learning and improved performance in single tasks? (Section \ref{section63})
    \item Does the representation encoded by TDRP effectively measure the transition distance? Are the representations robust to hyper-parameter selection? (Section \ref{section62})
    \item Can the auxiliary rewards produced by TDRP strengthen policy learning in skill-chaining tasks? (Section \ref{section64})
\end{itemize}
Section \ref{51} first introduces the experiment environments. Further experiment details are presented in Appendix \ref{appendixA}. The code is available in the supplementary material.

\subsection{Experiment Environments}
\label{51}
We conduct experiments in seven simulated robot manipulation tasks, as shown in Figure \ref{fig3}. 
Door-opening task (Door), Table-wiping task (Wipe), Nut-assembly task (NutAssemblySquare), and Two-arm-peg-in-hole task (TwoArmPegInHole) are from the Robosuite benchmark \cite{zhu2020robosuite}. In the Door-Opening task, the robot arm is trained to turn the handle and open doors. The Table-Wiping task requires the robot to clean the whiteboard surface and remove all markings. For the Nut-assembly task, the robot's objective is to correctly place a specific structured nut onto the matching peg. The Two-arm-peg-in-hole task presents a greater challenge, as it demands coordination between two robot arms to insert a peg from one arm into a hole in the other.
The Pick-nut (Pick), Place-nut-on-bolt (Place), and Screw-nut (Screw) tasks are from the Factory benchmark \cite{bib17}, which includes a Franka Panda arm, a table, a nut, and a bolt. All components have high-quality simulations and rich meshes, as the complex control required to solve the tasks demands high-fidelity simulations. 
All the tasks in Figure \ref{fig3} involve controlling one or two simulated robot arms to manipulate the target objects.

\begin{figure*}[htbp]
\centering
    \includegraphics[width=0.99\textwidth]{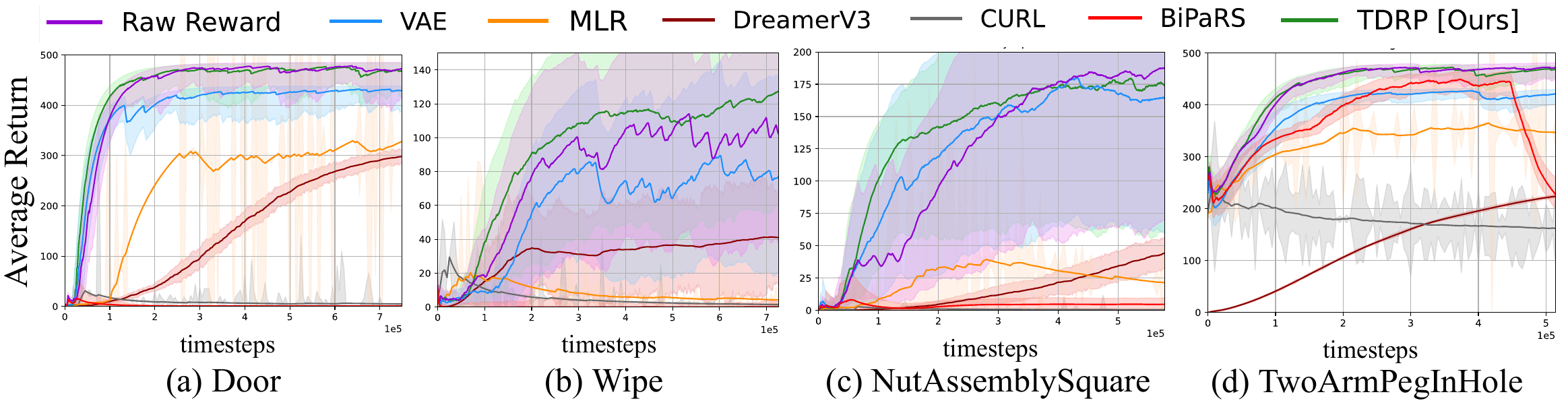}
\caption{Experiment results in the Robosuite tasks. The videos for the learned policies are provided in \url{https://sites.google.com/view/transition-distance-rp/tdrp}.}
\label{fig42}
\end{figure*}

\begin{figure*}[htbp]
\centering
    \includegraphics[width=0.7\textwidth]{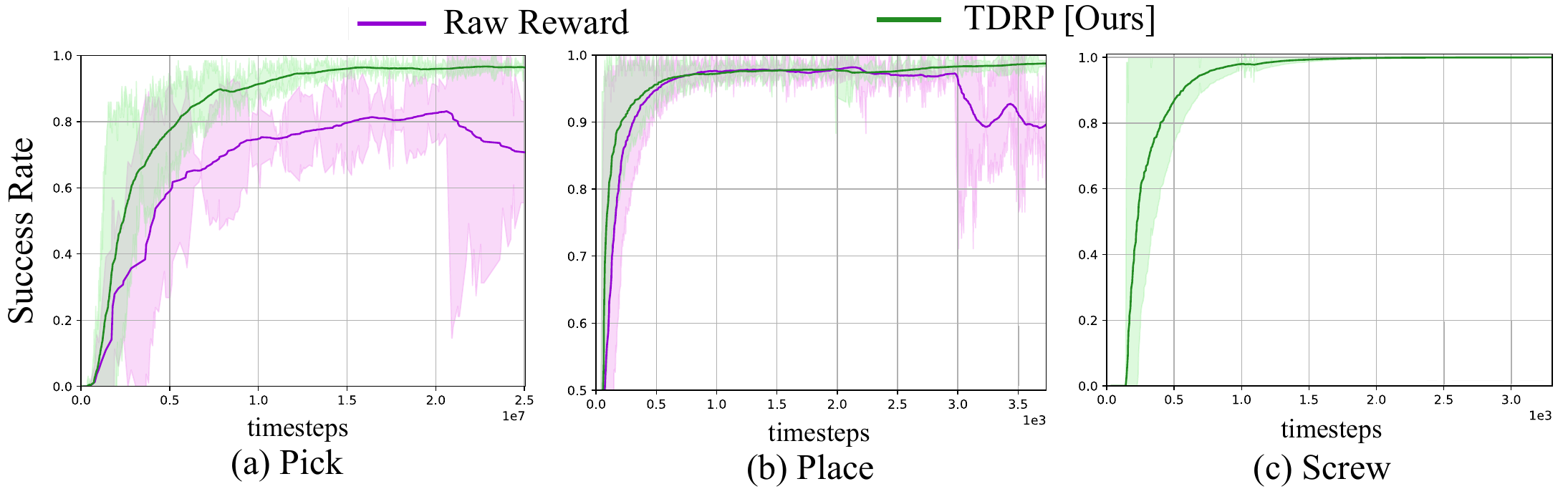}
\caption{Experiment results in the Factory tasks. (a) Pick-nut task. (b) Place-nut-on-bolt task. (c) Screw-nut task.}
\label{fig4}
\end{figure*}

\subsection{Experiment Results in Single Tasks} \label{section63}
We generate auxiliary rewards for the tasks in the Robosuite benchmark and train the policies with the reshaped rewards. 
As PPO is employed as the base RL algorithm, we compare optimizing the policy without auxiliary rewards using the PPO algorithm (denoted as ``Raw Reward'' in Figure \ref{fig42}).
Furthermore, we compare with generating auxiliary rewards in other representation spaces listed as follows, including reconstruction-based representations, model-based representations, and contrastive representations. Additionally, we compare our approach with the state-of-the-art reward shaping method.
\begin{itemize}
    \item VAE (Variational AutoEncoder) \cite{kingma2013auto}: We replace the TDRP encoder with VAE, and generate auxiliary rewards in the representation space learned by VAE\footnote{Note that the ways of generating rewards are the same for these methods, and the only difference is representation learning.}.
    \item MLR (Mask-based Latent Reconstruction for RL) \cite{yu2022mask}: A mask-based reconstruction method with representation prediction. 
    \item DreamerV3 (Mastering Diverse Domains through World Models) \cite{hafner2023mastering}: A general and scalable algorithm based on world models, where the representations are learned in a model-based way. 
    \item CURL (Contrastive unsupervised representations for RL) \cite{laskin2020curl}: A state-of-the-art contrastive representation approach with data augmentation.
    \item BiPaRS (Bi-level Parameterized Reward Shaping)\cite{2020Learning}: BiPaRS formulates reward shaping as a bi-level optimization problem, where the upper level centers on optimizing a parameterized shaping weight function for maximizing the true reward.
\end{itemize}

 The $y$ axis in Figure \ref{fig42} denotes the average return, where the return is accumulative for the raw rewards. Each curve is averaged over $5$ runs, and the shaded error bars represent the standard deviation.
 In the Door, NutAssemblySquare, and Wipe tasks, TDRP demonstrates better learning efficiency compared to the baseline methods. Notably, in the Wipe task, training with the proposed auxiliary rewards leads to a much larger convergent return. This phenomenon is possibly due to the TDRP's explicitly optimizing for a meaningful distance in the latent space, a characteristic absent in the baseline approaches.  The distance between the latent embeddings produced by TDRP serves as a measure of transition numbers between two states, which guides the agent to achieve more efficient learning.
 Due to the unstable bi-level structure in BiPaRS, the reward-shaping baseline BiPaRS can merely work in the easiest task, TwoArmPegInHole.

 \begin{table*}[htbp]
  \centering
  \fontsize{10pt}{14pt}\selectfont
  \resizebox{0.9\textwidth}{!}{
  \begin{tabular}{  c  c | c |c  c c c}
     &   \thead{Color-Timestep} & \thead{Raw state space (visualized by  tSNE\\ \cite{van2008visualizing})} & \multicolumn{4}{c}{Representations learned by TDRP}  \\ \hline
  & \multirow{2}{*}[28mm]{\adjustimage{width=0.37\textwidth, rotate=90}{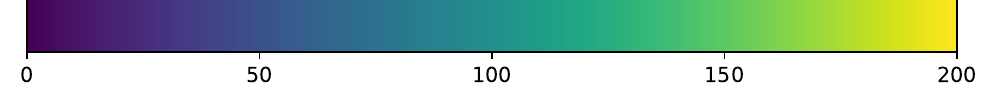}} &
    \multirow{2}{*}[10mm]{\fbox{\includegraphics[width=0.25\linewidth]{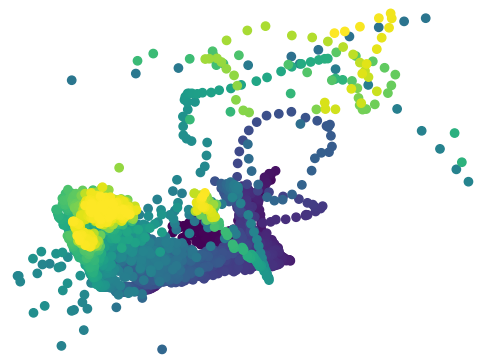}}} & 
    \rotatebox{90}{\ \ \ \ \ $step=10$} &
    \begin{minipage}[b]{0.5\columnwidth}
		\centering
		\includegraphics[width=\linewidth]{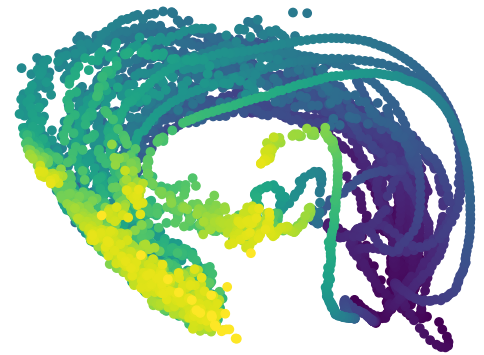}
	\end{minipage} &  \rotatebox{90}{\ \ \ \ \ $step=30$} &
    \begin{minipage}[b]{0.5\columnwidth}
		\centering
		\includegraphics[width=\linewidth]{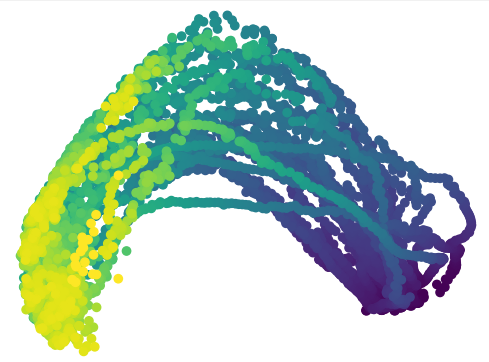}
	\end{minipage}   
    \\  ~& ~& ~& \rotatebox{90}{\ \ \ \ \ $step=50$} &
    \begin{minipage}[b]{0.5\columnwidth}
		\centering
		\includegraphics[width=\linewidth]{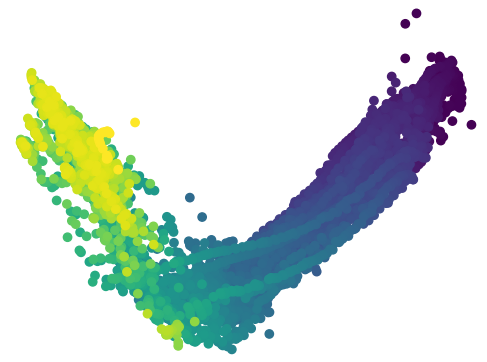}
    \end{minipage} &  \rotatebox{90}{\ \ \ \ \ $step=80$} &
    \begin{minipage}[b]{0.5\columnwidth}
		\centering
		\includegraphics[width=\linewidth]{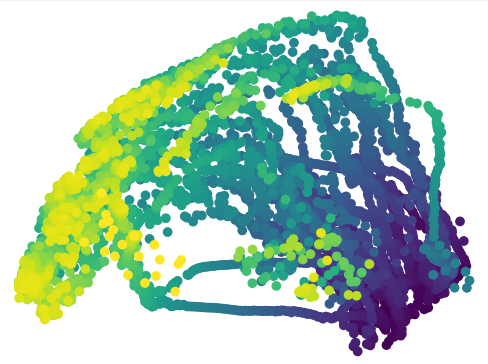}
    \end{minipage} 
  \end{tabular}
  }
  \caption{Visualization of the representations learned by the TDRP encoder in the Place-nut-on-bolt task. The colors of the points correspond to the timesteps of the states in the trajectory, as indicated by the ``Color-Timestep'' colorbar in the first column. }
  \label{table1}
\end{table*}

\subsection{Analysis of  the Learned Representations}\label{section62}
We further conduct experiments in the Factory benchmark to evaluate the generality of the proposed approach and inspect the learned latent representations. 
Figure \ref{fig4} compares the learning performance of using the proposed auxiliary rewards and without auxiliary rewards (``Raw Reward'').
We find that the learning of the proposed approach is more efficient, and converges to higher success rates.
Furthermore, training with the raw rewards tends to collapse over time, whereas training with our auxiliary rewards remains stable throughout. 
Especially in the Pick task, the convergent policy learned by our method significantly outperforms that learned with the raw rewards. 
In more complex tasks, such as Screw, using the raw environment rewards fails to learn successful policies (The success rate is always zero).

Table \ref{table1} visualizes the latent representations learned by TDRP in the Place task. 
The data for visualization are $30$ successful trajectories in this task.
Comparing the second column with the third column, we find that the proposed approach can reshape the unstructured trajectories into structured latent embeddings, where the Euclidean distance between two embeddings can measure their transition distance in the trajectories. This property contributes to the effective auxiliary rewards, which improves the success rates in Figure \ref{fig4}. We present the representation visualization in other tasks in Appendix \ref{appB}. Beyond that, we conduct an ablation study on the key hyper-parameter ``step'' in representation learning. The third column in Table \ref{table1} shows that TDRP can work with a large range of $step$, and with $step=30$ and $step=50$, the property of transition measurement is better. In the experiments of this paper, we set $step=50$.

\subsection{Experiment Results in Skill Chaining} \label{section64}
\begin{figure}[htbp]
\centering
    \includegraphics[width=0.5\textwidth]{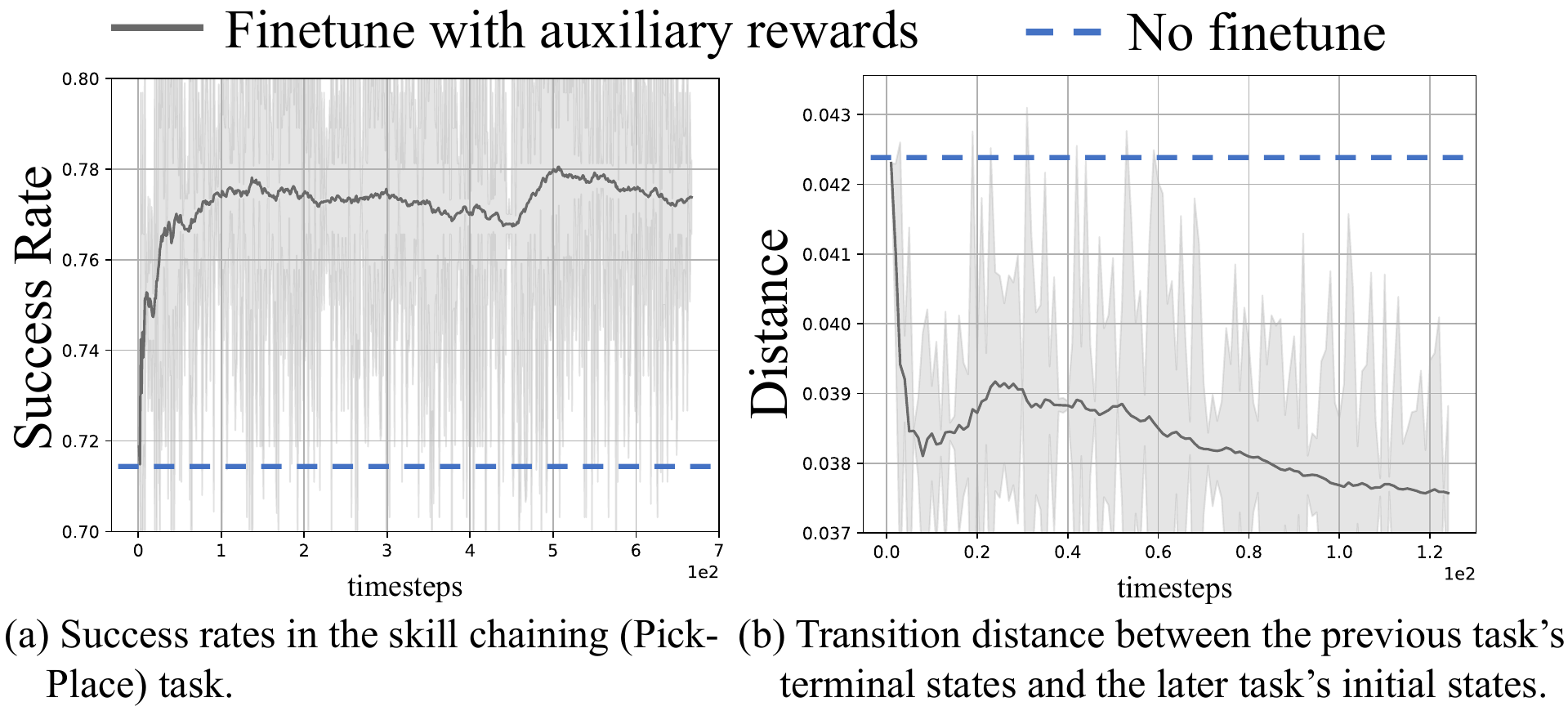}
\caption{Experiment results in the Pick-Place task.}
\label{fig7}
\end{figure}

 Skill chaining is effective for long-horizon challenging tasks, and in this subsection, we conduct experiments to figure out whether the proposed auxiliary rewards can improve skill-chaining performance.
 The target task is the Pick-Place task in the Factory benchmark, i.e., first picking the nut from the table and then placing the nut on the top of the bolt.
In Figure \ref{fig7}, we compare finetuning the skills with the auxiliary rewards in this long-horizon task with no finetuning.
Specifically, we extract the initial states of several successful trajectories of the Place task (latter skill) and use TDRP to encode these states. Then, we utilize the k-means method to generate $n$ clusters ($n=20$) and collect the center points of the $n$ clusters as the goal representation set for the Pick task (previous skill). 
Finally, we finetune the Pick skill with the proposed auxiliary rewards.

From Figure \ref{fig7}(a), we can see that the proposed auxiliary rewards can increase the success rate in the Pick-Place task by $6\%$. 
 In Figure \ref{fig7}(b), we visualize the transition distance between the terminal state of the Pick task and the initial state of the Place task during the fine-tuning process.  The transition distance decreased progressively, providing further evidence of our method's effectiveness.

\section{Conclusion and Future Work}\label{section7}
In this work, we introduce TDRP, an auxiliary reward generation approach with state representation learning. To generate effective auxiliary rewards for single tasks and skill chaining, TDRP extracts latent representations with the transition distance measurable property from the raw states. 
The experiments show that TDRP significantly improves the learning performance in single tasks and skill chaining, and substantially outperforms state-of-the-art baselines in RL. 
Beyond that, the representation visualization demonstrates that the learned representation can restructure the raw state space to a ``measurable'' latent space. 

Although TDRP is simple and broadly applicable, it has potential limitations in that the representation learning heavily depends on the online collected data. A large training data demand may harm the learning efficiency in challenging tasks with high-dimensional observation space. Generating auxiliary rewards in such challenging scenarios stands as a promising avenue for future work. In addition, for tasks with long horizons, careful selection of the goal states in the trajectories becomes crucial. In these scenarios, the integration of TDRP with hierarchical learning holds promise for extending the applicability of our method to more general tasks.

\bibliographystyle{named}
\bibliography{ijcai24}

\newpage

\clearpage

\appendix

\section{Experiment Details}\label{appendixA}
\begin{table*}[htbp]
    \centering
    \begin{tabular}{c|c c c }
        \hline\hline
         &Pick&Place&Screw\\ \hline
Action space dimension&23&23&23 \\
Episode length&100&200&1000 \\
Auxiliary Reward factory ($\lambda_1$)&2.5&3&0.02 \\ \hline
\multirow{2}{*}{Randomization}&Nut X-axis [-0.1, 0.1]m& Robot XY-axes [-0.2, 0.2]m&Nut X-axis [-0.0, 0.0]m\\
~&Nut Y-axis [-0.4, 0.2]m&Robot Z-axes [0.5,0.7]m&Nut Y-axis [-0.0, 0.0]m \\
\hline
    \end{tabular}
    \caption{Details on the Pick, Place, Screw tasks}
    \label{table2.1}
\end{table*}
\begin{table*}[]
    \centering
    \resizebox{\textwidth}{!}{
    \begin{tabular}{c|c c c c}
        \hline\hline
        &Door-opening &Table-wiping&Nut-assembly&Two-arm-peg-in-hole\\ \hline
Action space dimension & 12&12&12&12 \\
Episode length& 500&500&500&500 \\
Auxiliary reward scaling factor ($\lambda_1$)&0.05&0.01&0.005&0.0002 \\ \hline
\multirow{2}{*}{Randomization}&Door X-axis [0.07, 0.09]m&All position in table&Nut X-axis [0.11, 0.225]m& Robot XYZ-axes [-0.25, 0.25]m\\
&Door Y-axis [-0.01, 0.01]m~&~&Nut Y-axis [-0.225, -0.11]m& \\
\hline
    \end{tabular}
    }
    \caption{Details on the Door-opening, Wipe, NutAssemblySquare, TwoArmPegInHole tasks}
    \label{table2.2}
\end{table*}
\subsection{Task Descriptions}
\begin{itemize}
    \item \textbf{Pick-nut task} This task requires the robot arm to grasp the nut with a parallel-jaw gripper on a work surface.  The initial location of the nut is randomized in each episode.
    \item \textbf{Place-nut-on-bolt} In this scenario, a single robot arm transports the nut to the top of a bolt. The initial location of the robot is randomized and the bolt is fixed to the surface in each episode.
    \item \textbf{Screw-nut} This task requires screwing down the nut to a certain height, and the nut is initialized on the top of a bolt placed on a table.
    \item \textbf{Door-opening task} In this scenario, a door equipped with a handle is positioned in open space. A single robot arm learns how to turn the handle and open the door. The initial location of the door is randomized at the start of each episode.
    \item \textbf{Table-wiping task} In this setup, a table featuring a whiteboard surface with various markings is positioned in front of a single robot arm. The robot arm is equipped with a whiteboard eraser mounted on its hand, and its objective is to learn how to wipe the whiteboard surface clean, removing all of the markings. The initial arrangement of markings on the whiteboard is randomized at the start of each episode.
    \item \textbf{Nut-assembly task} In this scenario, a square peg is affixed to the tabletop. Additionally, a square nut is positioned in front of a single robot arm. The robot's objective is to correctly place the square nut onto the square peg. The initial placement of the nut is randomized at the commencement of each episode.
    \item \textbf{Two-arm-peg-in-hole task} In this setup, two robot arms are positioned side by side. One of the robot arms holds a board with a square hole in the center, while the other robot arm grasps a long peg. The task requires both robot arms to synchronize their actions to insert the peg into the hole. The initial configurations of the robot arms are randomized at the start of each episode.
    \item \textbf{Robot Arm: Panda} In all scenarios, the objective is to control one or two robot arms to accomplish the tasks. The Panda robot arm, a relatively new model manufactured by Franka Emika, features 7 degrees of freedom (7-DoF) and is known for its high positional accuracy and repeatability. The default gripper for this robot is the PandaGripper, a parallel-jaw gripper equipped with two small finger pads, which are included with the robot arm.
\end{itemize}
\subsection{Environment Details}
We evaluate the proposed approach on seven tasks from the Factory benchmark \cite{bib17} and the Robosuite benchmark \cite{zhu2020robosuite}.
The environment details are shown in Table \ref{table2.1} and \ref{table2.2}.

\subsection{Hyperparameter for the Baseline Algorithms}
When training the seven single tasks or the skill-chaining task using raw rewards, the auxiliary rewards generated by our method (TDRP [Ours]), and VAE (variational autoencoders) \cite{kingma2013auto}, we employ the Proximal Policy Optimization algorithm (PPO) \cite{schulman2017proximal} as the reinforcement learning (RL) algorithm. We list the hyper-parameter of PPO in Table \ref{table3}.
\begin{table}[]
    \centering
    \begin{tabular}{c|c}
        \hline \hline
         \textbf{Parameter}&\textbf{Value}  \\ \hline 
          MLP network size&[256,128,64] \\
          Length $H$ &32\\
          Adam learning rate&1.0e-4\\
          Discount factory&0.99\\
          GAE parameter&0.95\\
          Entropy coefficient&0.0\\
          Critic coefficient&2.0\\
          Minibatch size&512\\
          Minibatch epochs&8\\
          Clipping parameter&0.2\\ \hline
    \end{tabular}
    \caption{Hyperparameters of PPO}
    \label{table3}
\end{table}

The hyperparameters of the three baseline algorithms: CURL (Contrastive unsupervised representations for RL), DreamerV3 (Mastering Diverse Domains through World Models), and MLR (Mask-based Latent Reconstruction for RL) are listed in Table \ref{table4}. 

\begin{table*}[htbp]
    \centering
    \begin{tabular}{c|c|c}
        \hline \hline
         \textbf{Algorithms}&\textbf{Parameter}&\textbf{Value}  \\ \hline 
          \multirow{11}{*}{\textbf{CURL}}&Q network: hidden units&256 \\
          &Momentum&0.001\\
          &Non-linearity&ReLU\\
          &Update&Distributional Double Q\\
          &Discount factory&0.99\\
          &Batch size&32\\
          &Optimizer&Adam\\
          &Optimizer: learning rate&0.0001\\
          &Optimizer: $\theta_1$&0.9\\
          &Optimizer: $\theta_2$&0.999\\
          &Optimizer $\epsilon$&0.000015\\ \hline
        \multirow{10}{*}{\textbf{DreamerV3}}&Non-linearity&SiLU\\
        &Batch size&16 \\
        &Batch length&64\\
        &Reconstruction loss scale&1.0\\
        &Representation loss scale&0.1\\
        &Critic EMA decay&0.98\\
        &Critic EMA regularizer&1\\
        &Learning rate&$3\times10^{-5}$\\
        &Adam epsilon&$10^{-5}$\\
        &Gradient clipping&100\\ \hline
        \multirow{8}{*}{\textbf{MLR}}&Optimizer&Adam\\
        &Learning rate&0.0001\\
        &Batch size for policy learning&512\\
        &Batch size for auxiliary task&128\\
        &Discount factor&0.99\\
        &Weight of MLR loss&1\\
        &Mini-batch size&32\\
        &Mask ratio&50\%\\ \hline
    \end{tabular}
    \caption{Hyperparameters of baselines}
    \label{table4}
\end{table*}

\section{Additional Experiment Results}\label{appB}

In Table \ref{table_pick}, we visualize the representations learned by the TDRP encoder in the Pick-nut task. The colors of the points correspond to the timesteps of the states in the trajectory, represented by the ``Color-Timestep'' colorbar in the first column.
The results demonstrate that the proposed approach can roughly learn the representations with a transition measure property, and is robust to the hyperparameter $step$.

 \begin{table*}[htbp]
  \centering
  \fontsize{10pt}{14pt}\selectfont
  \resizebox{0.9\textwidth}{!}{
  \begin{tabular}{  c  c | c |c  c c c}
     &   \thead{Color-Timestep} & \thead{Raw state space (visualized by  tSNE\\ \cite{van2008visualizing})} & \multicolumn{4}{c}{Representations learned by TDRP}  \\ \hline
  & \multirow{2}{*}[28mm]{\adjustimage{width=0.37\textwidth, rotate=90}{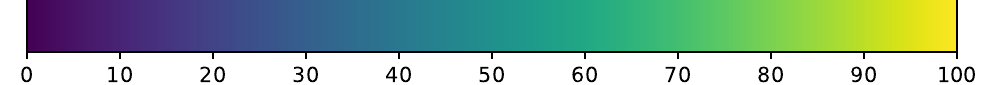}} &
    \multirow{2}{*}[10mm]{\fbox{\includegraphics[width=0.25\linewidth]{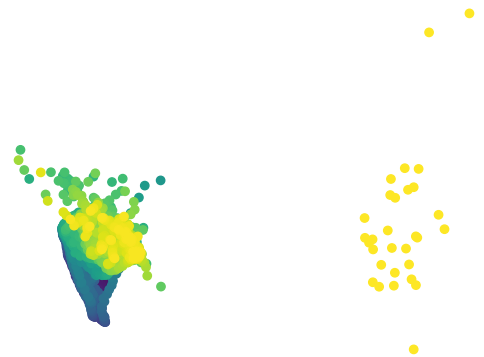}}} & 
    \rotatebox{90}{\ \ \ \ \ $step=10$} &
    \begin{minipage}[b]{0.5\columnwidth}
		\centering
		\includegraphics[width=\linewidth]{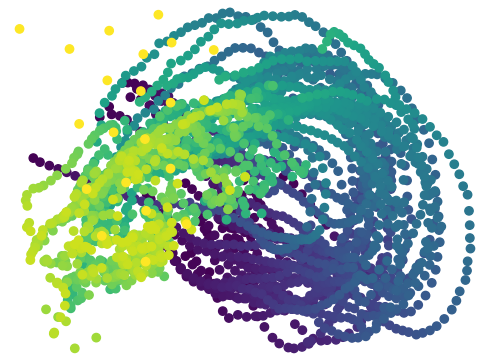}
	\end{minipage} &  \rotatebox{90}{\ \ \ \ \ $step=30$} &
    \begin{minipage}[b]{0.5\columnwidth}
		\centering
		\includegraphics[width=\linewidth]{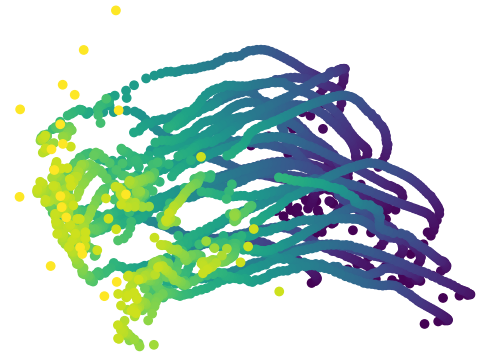}
	\end{minipage}   
    \\  ~& ~& ~& \rotatebox{90}{\ \ \ \ \ $step=50$} &
    \begin{minipage}[b]{0.5\columnwidth}
		\centering
		\includegraphics[width=\linewidth]{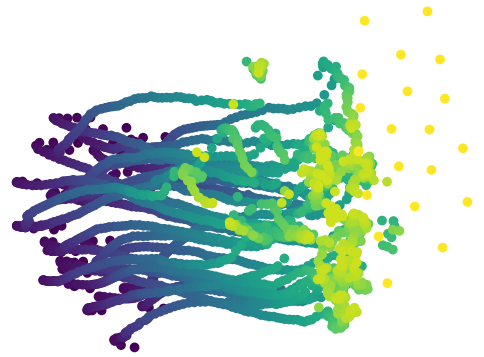}
    \end{minipage} &  \rotatebox{90}{\ \ \ \ \ $step=80$} &
    \begin{minipage}[b]{0.5\columnwidth}
		\centering
		\includegraphics[width=\linewidth]{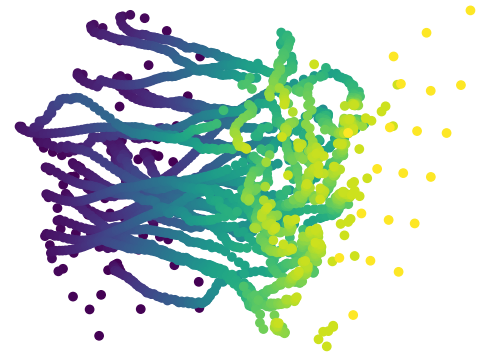}
    \end{minipage} 
  \end{tabular}
  }
  \caption{Visualization of the representations learned by the TDRP encoder in the Pick-nut task. }
  \label{table_pick}
\end{table*}






\appendix
\end{document}